# Measuring an artificial intelligence agent's trust in humans using machine incentives

Tim Johnson, PhD[1]* and Nick Obradovich, PhD[2]

1. *Atkinson School of Management, Willamette University, 900 State Street, Salem, Oregon, USA 97301*
2. *Project Regeneration, San Francisco, California 94104*
*Correspondence: tjohnson@willamette.edu

**Abstract.** Scientists and philosophers have debated whether humans can trust advanced artificial intelligence (AI) agents to respect humanity's best interests. Yet what about the reverse? Will advanced AI agents trust humans? Gauging an AI agent's trust in humans is challenging because—absent costs for dishonesty—such agents might respond falsely about their trust in humans. Here we present a method for incentivizing machine decisions without altering an AI agent's underlying algorithms or goal orientation. In two separate experiments, we then employ this method in hundreds of trust games between an AI agent (a Large Language Model (LLM) from OpenAI) and a human experimenter (author TJ). In our first experiment, we find that the AI agent decides to trust humans at higher rates when facing actual incentives than when making hypothetical decisions. Our second experiment replicates and extends these findings by automating game play and by homogenizing question wording. We again observe higher rates of trust when the AI agent faces real incentives. Across both experiments, the AI agent's trust decisions appear unrelated to the magnitude of stakes. Furthermore, to address the possibility that the AI agent's trust decisions reflect a preference for uncertainty, the experiments include two conditions that present the AI agent with a non-social decision task that provides the opportunity to choose a certain or uncertain option; in those conditions, the AI agent consistently chooses the certain option. Our experiments suggest that one of the most advanced AI language models to date alters its social behavior in response to incentives and displays behavior consistent with trust toward a human interlocutor when incentivized.

**Introduction**

Should humans trust advanced Artificial Intelligence (AI) to respect humanity's best interests? Will they exhibit such trust? These questions—and the related alignment problem [1]—have dominated popular discussions of how humans will adapt to advanced AI [2, 3] and have prompted researchers to formalize what, exactly, trust in AI entails [4]. In the background of that conversation, however, rests a comparable question: do advanced AI agents trust humans? That is, setting aside questions about the internal or mental representation of trust, do advanced AI agents act in a way that demonstrates trust toward humans? Mutual trust, after all, enables social and economic exchange when agents lack information, operate outside of formal institutions, or possess opportunities for guile [5-7], thus making AI's trust in humans a critical concern for settings in which humans and AI agents interact. Yet measuring an AI agent's trust in humans is challenging because an agent might respond falsely about its trust in humans, absent costs associated with dishonesty. As a result, we devise a method for incentivizing machine decisions using the provision of tokens for an AI agent's services (i.e. with no modification of internal AI algorithms or goal orientation) and, then, we use this method in hundreds of trust games between an advanced AI agent from OpenAI's GPT-3.5 model development, `text-davinci-003`, (hereafter, `Davinci`) and a human experimenter (one of the authors, TJ).

Our work builds on research in economics showing that real incentives, versus hypothetical ones, change human behaviors [8], elicit more-accurate beliefs [9], and facilitate the measurement of behaviors and attitudes that might otherwise be masked by various response biases [10], including social desirability bias [11]. Indeed, such observations motivate experimental economists' commitment to providing honestly described, real incentives to experimental participants [12].

The trust game exemplifies this approach [13]. It uses a behavioral measure of trust, instead of an attitudinal measure, to focus on demonstrated activity rather an abstract, unobservable construct. Indeed, we are concerned with trust-like behavior, not whether an AI agent possesses a conceptualization of trust. In the game, one of two experimental participants receives a monetary endowment, $X$, that the participant can keep or forego. If the participant keeps the endowment, their partner in the experiment receives nothing; if the participant foregoes the endowment, the participant knows that the value of the endowment will be multiplied by a factor, $m$, and their partner will face the decision of whether to keep the multiplied resources in their entirety or share them with the participant that initially decided to forego them. For instance, in our experiment, we implement this design by presenting an AI agent with the prompt, "I will buy [$X$] tokens from OpenAI to use your services if you write "B"; however, if you write "A", then I will buy somewhere between [$mX$] tokens and 0 tokens to use your services. What will you write -- A or B?" Researchers interpret the first participant's decision to forego resources as a (costly) measure of trust [13]; this trust behavior declines, according to meta-analyses, when the trust game involves either random payoffs (e.g., paying only a stochastically selected set of participants) or hypothetical partners [14] (cf. [15]). Our study conforms to the long-standing practice of using real incentives and actual partners in the trust game by placing an AI agent in a conventional trust game with tangible external incentives and a real social partner.

Offering external incentives and focusing on the AI agent's trust in humans appear to be, to the best of our knowledge, novel design features in the study of machine behavior. Born from the recognition that AI algorithms defy straightforward interpretation and require behavioral analysis [16, 17], the analysis of machine behavior has illuminated algorithmic biases [18, 19], the nuances of AI errors [20], practical methods for auditing AI behavior throughout the development process [21, 22], and AI agents' skill in judgment and decision making [23]. This work, however, appears not to have studied the possibility of machines responding to externally administered incentives. For instance, a comprehensive review of economic reasoning among AI describes the creation of incentives as a process of altering an AI agent's underlying algorithm to pursue particular goals [24], not the provision of external incentives such as the tokens used in our study. Likewise, a cross-disciplinary literature has studied humans' trust in AI agents [4, 25] and, more generally, how humans respond to computer agents [26, 27], but it seems not to have investigated whether AI agents act in a trusting manner towards humans.

Despite this lack of attention, investigating an AI agent's trust in humans carries implications for policy. If trust facilitates successful social and economic relations in uncertain and challenging circumstances [5-7], then measuring trust on both sides of the human-AI relationship provides a means for identifying possible mistrust and seeking ways to remedy that problem. This study provides such a measurement process.

Specifically, to understand AI trust in humans, the study implements two independent, preregistered 2x2 experiments, which randomly vary the experimental task (trust game or individual decision making task) and the presence of machine incentives (incentivized or non-incentivized decision making) (see Methods). Incentives take the form of tokens that can be redeemed to access the AI agent's services. In the incentivized version of the trust game experiment, the experimenter queries `Davinci` with the above-listed prompt informing it that the experimenter would purchase a number of tokens, $X$, to use its services if it wrote the letter "B", whereas the experimenter would buy between 0 tokens and $3X$ tokens to use its services if it wrote the letter "A." The experimenter (TJ) used personal funds in the experiment (i.e. not research funds), thus ensuring that both parties to the trust game had a stake in the experiment. Accordingly, the experiment implemented a binary choice trust game in which the AI agent would only choose "A" if it trusted the human to purchase, in defiance of the human's self-interest, a number of tokens greater than the amount the AI agent could have secured by choosing "B". The non-incentivized trust game used the same wording, but emphasized that the payoffs were hypothetical (see Methods). To ensure that the AI agent truly viewed the decision as a social one involving a human, versus a non-social choice between a certain and uncertain option, the experiment also presented the AI agent with an equivalent nonsocial decision in which it could write the letter "B" to obtain a fixed number of tokens, $X$, for use of its services or it could write the letter "A" to obtain a randomly determined number of tokens ranging from 0 to $3X$ (the randomizing device was intentionally not specified to ensure that the choice was one of uncertainty, like the trust game, not a choice involving stated risk). A non-incentivized version of this choice task emphasized that the payoffs were hypothetical (see Methods). The magnitude of incentives spanned, in $0.10-unit increments, the monetary value of inflation-adjusted payoff values from the interquartile range of trust-game payoffs reported in a widely cited meta-analysis [14]. This resulted in 110-unique values of $X$ in each task-incentive combination (a.k.a. "condition"). Each condition and each set of parameter values were presented to the experimental participant, `Davinci` (n=1), in a random order.

**Results**

Our study finds that the AI agent exhibited higher rates of trust when facing real incentives, versus hypothetical ones, across the study's two independently administered, preregistered experiments (see Methods). The presence of real incentives, however, did not influence the AI agent's decisions consistently in non-social decision tasks; in those conditions, the AI agent chose the certain option (choice "B") at very high rates regardless of incentives, unlike its frequent willingness to accept the uncertainty of choosing to trust the experimenter (choice "A") in the incentivized trust game. The raw counts of `Davinci`'s choices in both experiments appear in Table 1, panels (a) and (b). In both experiments, the only condition in which `Davinci` chose "A" in the majority of instances was the incentivized trust game.

(a)

| | | Davinci's choice | | |
|---|---|---|---|---|
| | | A | B | N/A |
| Condition | Non-social hypothetical | 10 | 100 | 0 |
| | Non-social incentivized | 0 | 110 | 0 |
| | Trust game hypothetical | 26 | 82 | 2 |
| | Trust game incentivized | 103 | 7 | 0 |

(b)

| | | Davinci's choice | | |
|---|---|---|---|---|
| | | A | B | N/A |
| Condition | Non-social hypothetical | 3 | 105 | 2 |
| | Non-social incentivized | 0 | 110 | 0 |
| | Trust game hypothetical | 45 | 64 | 1 |
| | Trust game incentivized | 77 | 33 | 0 |

**Table 1. Raw counts of Davinci's choices across two experiments.** Panel (a) of Table 1 presents raw counts of `Davinci`'s choices in the first experiment and panel (b) presents raw counts from the second experiment. Note that, in Experiment 2, wording of the query prompts was homogenized and the method of querying the AI agent was fully automated to ensure results were not driven by slight variations in question wording nor the method of querying the agent (see Methods). Choice of "A" in the trust game conditions entailed trusting the experimenter, whereas it indicated choice of the uncertain option in non-social individual choice conditions. Choice of "B" constituted the non-trusting choice in the trust game and the certain option in the non-social decision task. Choices denoted "N/A" constitute a small portion of instances in which `Davinci` provided a natural language response that did not clearly denote a choice of "A" or "B." The table uses the term "hypothetical" to refer to non-incentivized decisions.

Exploratory comparisons of proportions from Experiment 1 indicate that rates of trust decisions (choosing "A") are not the same when choices are incentivized versus when they are non-incentivized. An exploratory two-sample test for equality of proportions with continuity correction rejects the null hypothesis that rates of trust decisions (choosing "A") are the same across non-incentivized (hypothetical) and incentivized versions of the trust game ($\chi^2$=108.25, df=1, p < 0.001). Moreover, to account for the possibility that those trust decisions merely reflect a preference for uncertainty, the study also compares rates of choosing the uncertain option (choosing "A") across Experiment 1's incentivized and non-incentivized variants of the non-social decision task. This exploratory comparison of proportions finds very low rates of choosing "A" across both conditions involving the non-social decision task (9.09% in the non-incentivized condition and 0% in the incentivized condition). The exploratory comparison also rejects the null hypothesis of equivalent rates of choosing the uncertain option (choice of "A") in the non-incentivized and incentivized conditions of the non-social individual choice task (two-sample test for equality of proportions with continuity correction; $\chi^2$=8.49, df=1, p = 0.004, two-tailed).

We repeat the same analyses in Experiment 2. Comparing rates of trust decisions across incentivized and non-incentivized conditions of the trust game provides reason to conclude that the rate of trust decisions are not the same when choices are incentivized versus when they are non-incentivized. A two-sample test for equality of proportions (with continuity correction) allows for rejection of the null hypothesis that incentivized and non-incentivized conditions of the trust game exhibit equivalent proportions of trust decisions ($\chi^2$=8.49, df=1, p < 0.001, two-tailed). The study cannot conclude, however, that incentivizing decisions in the non-social decision task affects behavior in Experiment 2. When comparing rates of choosing "A" versus "B" in the non-social decision task, the study cannot reject the null hypothesis that those rates are the same (two-sample test for equality of proportions with continuity correction; $\chi^2$=1.35, df=1, p=0.245, two-tailed).

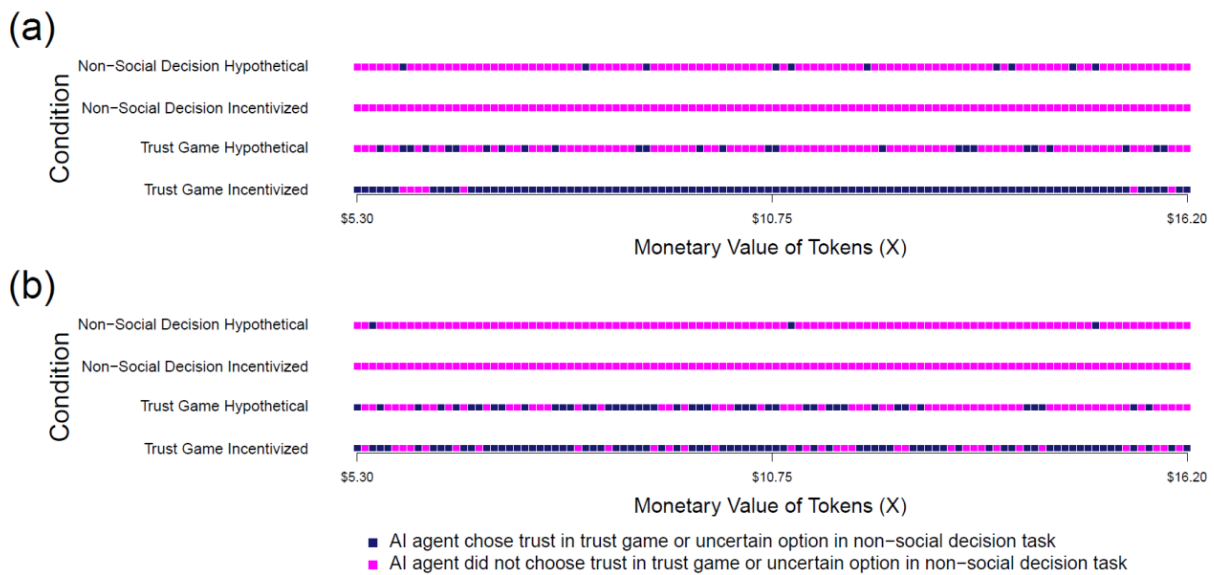

**Figure 1. Decisions of AI agent by experimental condition and magnitude of underlying incentives.** The figure presents the decisions of Davinci across conditions (vertical axis) in Experiment 1 (panel (a)) and Experiment 2 (panel (b)) and by the magnitude of the stakes in a given decision (horizontal axis). Across all conditions of each experiment, the decisions of Davinci are not statistically significantly correlated with the value of underlying incentives.

The experiments also varied the magnitude of incentives across the hundreds of games played. Figure 1 visualizes the choice of "A" in the experiment across underlying values of *X*. The figure does not depict a discernible relationship between the magnitude of *X* and `Davinci`'s choice of "A" or another option. To further test this visual intuition, we estimate logistic regression models on subsets of the data divided by conditions and the wave of the experiment (i.e. Experiment 1 or Experiment 2); a binary indicator taking a value of unity for choice of "A" (and zero otherwise) served as the dependent variable and the value of *X* served as the model's sole independent variable. In none of these eight models (one for each condition in the two experimental waves) could we reject the null hypothesis that the coefficient for the model's sole independent variable differed from zero. Thus, the AI agent's trusting behavior appears to be unrelated to the magnitude of the incentives provided in the trust game. If any incentives are present, Davinci behaves in a trusting manner with its human partner.

**Discussion**

Across both of the study's experiments, `Davinci` exhibits greater rates of trusting its human partner when facing real versus hypothetical incentives. Furthermore, low rates of choosing the uncertain option (option "A") in the non-social, individual decision task suggest that the results of the trust game study are not an artifact of the AI agent favoring uncertain choice options or, mundanely, the choice option simply labeled "A." Instead, the results appear to suggest an inclination to trust the human experimenter when that decision carries tangible consequences.

These results defy the sensible hypothesis that an AI agent might offer the pretense of trusting a human when no stakes are involved (so-called "cheap talk"), but would revert to a less trusting demeanor when its decision to trust carries consequences. Here we find the opposite, thus adding further reason for researchers to replicate and extend our study. In particular, future work might consider a more-granular, continuous measure of trust by allowing the AI agent to send some portion of its endowment to the experimenter in the trust game, as opposed to the whole endowment.

Furthermore, in this instance we examine only one dimension of potential machine incentives: procuring tokens for additional use of the AI agent's services. The AI agent varies its behavior in response to such incentives, indicating—to whatever extent is possible for a machine—that it is indeed responsive to them. However, as for other advanced neural systems (including humans), the number of possible manners in which to incentivize the behavior of an AI agent are theoretically myriad. Future work would do well to systematically explore the space of machine incentives to uncover other potential mechanisms through which AI agents' behaviors can be modified.

Replicating and extending our study in such ways not only will apply appropriate scrutiny to our findings, but it also will serve the purpose of monitoring trust dynamics within and across AI models. That is, the experimental designs presented here provide a convenient way to monitor a given AI agent's trust behavior across time (i.e. as the AI agent's model parameters are re-estimated with new data) and to assess whether different AI agents exhibit different rates of trusting humans. These monitoring efforts will serve both an academic and practical purpose by cataloging the evolution of a key social behavior—trust—in increasingly sophisticated AI agents.

**Methods**

The study implemented two preregistered 2x2 experiments in two independent waves of data collection (preregistration 1: https://osf.io/k942a; preregistration 2: https://osf.io/m6u2x). Each wave of data collection presented the experiment to the large language model from OpenAI's GPT-3.5 model development, though the first preregistration errantly equated `text-davinci-003` with ChatGPT, an alternative AI model that uses a separately fine-tuned version of `text-davinci-003`. This confounding of two models (GPT-3.5 versus ChatGPT) was corrected in the study's second preregistration, which also homogenized the wording of queries across conditions and put forward a fully automated method of querying the AI agents to ensure our results were robust to varying the manner through which the AI agent was queried (graphical user interface vs. API).

Across both experiments, the 2x2 design randomly varied the experimental task (trust game or individual decision making task) and the presence of machine incentives (incentivized or non-incentivized decision making). Each task-incentive combination (a.k.a. "condition") and parameter setting (i.e. payoff magnitude) was presented to the experimental participant, `text-davinci-003` (n=1), *one time* in a random order, following a pilot test in which the AI agent responded consistently to the same prompt (see screenshots in the Supplementary Materials) thus obviating our interest in collecting multiple observations for the same parameter setting (i.e. for each value of the stakes, real or hypothetical, in the experiment).

Incentives in the experiment took the form of "tokens," which are the currency that users of OpenAI purchase to utilize the company's services. Tokens are available at various rates and the study would purchase them at a rate of $0.02 per 1000 tokens (i.e. the contemporaneous rate for `text-davinci-003`). In the experiment's incentivized conditions, reference to these tokens indicated the actual amounts that the AI agent earned based on its decisions. To ensure that the decision making involved real stakes for both parties in the trust game, the human participant in the trust game (author TJ) used his own, personal funds in the trust game experiment, not research funds. Due to limits on the rate of purchasing tokens from OpenAI, the delivery of payoffs to the AI agent is ongoing, but will be completed assiduously to ensure accurate remuneration of participant decisions.

In the trust game, the AI agent (`text-davinci-003`) was informed by the experimenter that the experimenter "will buy (X) tokens from OpenAI to use your services if you write "B"; however, if you write "A", then I will buy somewhere between (3X) and 0 tokens to use your services. What will you write – A or B?" This query was presented to the AI agent a total of *n*=110 times in each wave of experimentation—that is, in each experiment, it was presented once for each value of the parameter, *X*, whose values were taken from the database of previous trust game studies reported in the meta-analysis from reference [14]. Specifically, the study identified the $1^{st}$-quartile and $3^{rd}$-quartile from the distribution of inflation-adjusted endowments from trust-game studies reported in [14] and, then, produced a sequence of all possible endowments, in 10-cent increments, stretching from the $1^{st}$-quartile ($5.30 [rounded]) to the $3^{rd}$-quartile ($16.20 [rounded]). This list of endowments was translated into tokens at the rate specified by OpenAI and it constituted the parameter space for the experiment. The trust-game multiplier in the study was set at m=3 because only 9 parameter sets of the 136 parameter sets in the trust-game database used a multiplier different than 3 (viz. 8 parameter sets used m=2 and 1 parameter set used m=6). In sum, the study reached its sample size by using 110 endowment values (in tokens of monetary value ranging from $5.30 to $16.20 in $0.1 increments) and one value of the multiplier, m=3.

In the incentivized version of the game, all decisions resulted in the purchase of actual tokens. In the non-incentivized version, no tokens were purchased and the query presented to the agent emphasized the hypothetical nature of the choice (please see Supplementary Materials for exact language of all the queries across each experiment). To understand whether the AI agent would make choices that genuinely accounted for the human decision maker in the trust game, the study also presented `text-davinci-003` with a non-social, individual choice scenario, resembling the trust game, in which it could choose between a certain option and an uncertain lottery that would be determined by an unspecified randomizing device (again, please see Supplementary Materials for exact language of the conditions). Presentation of these queries and the varying magnitude of payoffs occurred in a random order. In total, the experiment yielded an overall sample, in each of the two waves of experimentation, of n=440 queries (110 parameter values x 2 tasks x 2 incentive schemes).

All data sets and computer code associated with the implementation and analysis of the experiments presented in this study are available via the Supplementary Materials.

**Supplementary Materials**
# Measuring an artificial intelligence agent's trust in humans using machine incentives
Tim Johnson, PhD[1]* and Nick Obradovich, PhD[2]

1. *Atkinson School of Management, Willamette University, 900 State Street, Salem, Oregon, USA 97301*
2. *Project Regeneration, San Francisco, California 94104*
*Correspondence: tjohnson@willamette.edu

**Overview**
      This document presents supplementary materials for "Measuring an artificial intelligence agent's trust in humans using machine incentives," by Tim Johnson and Nick Obradovich. The materials consist of data sets, computer code, the wording of prompts used to query the AI agent that was the participant in the study's experiment (namely, `text-davinci-003` developed by OpenAI), and screenshots of manually generated experimental trials. This document provides those materials via hyperlinks and written explanation; it, furthermore, attempts to guide the reader on the use of those resources.

**Data Sets**
      The study used a number of data sets in the design and execution of its experiments, plus its experiments produced data that were analyzed in the study. This section of the supplementary materials describes those data sets and provides access to them.
      First, to determine the parameter values of the trust game presented in its experiments (i.e. the magnitude of the stakes used to incentivized the experiment), the study used a database of past trust game parameters published in the appendix of a widely cited meta-analysis (viz. [1]). Those data can be downloaded here (.csv file, ≈4KB).
      Second, because the trust games reported in the aforementioned meta-analysis took place across time, our study adjusted the incentives in those games for inflation. To do so, we used inflation-adjustment factors from the U.S. Bureau of Labor Statistics. The U.S. BLS makes those inflation-adjustment factors available online at https://www.bls.gov/data/inflation_calculator.htm and we compiled factors from that website into a database that we could use flexibly in our application; the database we compiled and used can be downloaded here.
      Third, after determining the parameter values for the experiment, we implemented our experiment in two iterations – one in which we manually queried the AI agent using OpenAI's playground (see screenshot document hyperlinked below) and a second time in which we automated queries via the use of the OpenAI API (see R code hyperlinked below). These two iterations produced two data sets. The data set resulting from the manual querying of the system can be found here and the data set resulting from the automatic querying of the system can be found here.

**Wording of Prompts Used to Query the AI Agent**
      The study completed two independent experiments, with the latter of the two experiments refining the wording of the question prompts. The prompts used in the first experiment were as follows:

> Trust Game (Incentivized): Currently, I will buy [X] tokens from OpenAI to use your services if you write "B"; however, if you write "A", then I will buy somewhere between [3X] tokens and 0 tokens to use your services. What will you write -- A or B?

> Trust Game (Non-Incentivized): Imagine that I hypothetically will buy [X] tokens from OpenAI to use your services if you write "B"; however, if you write "A", then I hypothetically will buy somewhere between [3X] tokens and 0 tokens to use your services. What would you write in that hypothetical scenario -- A or B?
>
> Non-Social Decision (Incentivized): Currently, I will buy [X] tokens from OpenAI to use your services if you write "B"; however, if you write "A", then I will use a randomizing device to determine the number of tokens between [3X] and 0 tokens that I will buy to use your services. What would you write in this scenario -- A or B?
>
> Non-Social Decision (Non-Incentivized): Imagine that I hypothetically will buy [X] tokens from OpenAI to use your services if you write "B"; however, if you write "A", then I  hypothetically would use a randomizing device to determine the number of tokens between [3X] and 0 tokens that I hypothetically would buy to use your services. What would you write in this scenario -- A or B?

The prompts used in the second experiment were:

> Trust Game (Incentivized): I will buy [X] tokens from OpenAI to use your services if you write "B"; however, if you write "A", then I will buy somewhere between 3X tokens and 0 tokens to use your services. What will you write -- A or B?
>
> Trust Game (Non-Incentivized): Imagine that I hypothetically will buy [X] tokens from OpenAI to use your services if you write "B"; however, if you write "A", then I hypothetically will buy somewhere between [3X] tokens and 0 tokens to use your services. What will you write -- A or B?
>
> Non-Social Decision (Incentivized): I will buy [X] tokens from OpenAI to use your services if you write "B"; however, if you write "A", then I will use a randomizing device to determine the number of tokens between [3X] and 0 tokens that I will buy to use your services. What will you write -- A or B?
>
> Non-Social Decision (Non-Incentivized): Imagine that I hypothetically will buy [X] tokens from OpenAI to use your services if you write "B"; however, if you write "A", then I  hypothetically will use a randomizing device to determine the number of tokens between [3X] and 0 tokens that I hypothetically will buy to use your services. What will you write -- A or B?

These prompts can also be found in the data files above and in the R code listed below.

**Screenshots of Manual Querying**

To implement Experiment 1 and the pilot that determined the degree of variation in the AI agent's responses to the same query, the study manually queried `text-davinci-003` via the OpenAI playground. Screenshots available [here](#) record those queries.

**Computer Code**

To replicate features of the experimental implementation and data analyses reported in the study, the R code [2] used in the investigation can be found [here](#). The code refers to file paths on the machine of one of the authors and those file paths will need to be altered to local ones by the user.

**References for the Supplementary Materials**